# An Improved Search Algorithm
# for Optimal Multiple-Sequence Alignment

**Stefan Schroedl**                                        STEFAN.SCHROEDL@GMX.DE
*848 14th St*
*San Francisco CA 94114*
*+1 (415) 522-1148*

## Abstract

*Multiple sequence alignment (MSA)* is a ubiquitous problem in computational biology. Although it is *NP*-hard to find an *optimal* solution for an arbitrary number of sequences, due to the importance of this problem researchers are trying to push the limits of exact algorithms further. Since MSA can be cast as a classical path finding problem, it is attracting a growing number of AI researchers interested in heuristic search algorithms as a challenge with actual practical relevance.

In this paper, we first review two previous, complementary lines of research. Based on *Hirschberg's algorithm, Dynamic Programming* needs $O(kN^{k-1})$ space to store both the search frontier and the nodes needed to reconstruct the solution path, for $k$ sequences of length $N$. *Best first search*, on the other hand, has the advantage of bounding the search space that has to be explored using a heuristic. However, it is necessary to maintain all explored nodes up to the final solution in order to prevent the search from re-expanding them at higher cost. Earlier approaches to reduce the *Closed* list are either incompatible with pruning methods for the *Open* list, or must retain at least the boundary of the *Closed* list.

In this article, we present an algorithm that attempts at combining the respective advantages; like $A^*$ it uses a heuristic for pruning the search space, but reduces both the maximum *Open* and *Closed* size to $O(kN^{k-1})$, as in Dynamic Programming. The underlying idea is to conduct a series of searches with successively increasing upper bounds, but using the DP ordering as the key for the *Open* priority queue. With a suitable choice of thresholds, in practice, a running time below four times that of $A^*$ can be expected.

In our experiments we show that our algorithm outperforms one of the currently most successful algorithms for optimal multiple sequence alignments, *Partial Expansion $A^*$*, both in time and memory. Moreover, we apply a *refined heuristic* based on optimal alignments not only of pairs of sequences, but of larger subsets. This idea is not new; however, to make it practically relevant we show that it is equally important to bound the heuristic computation appropriately, or the overhead can obliterate any possible gain.

Furthermore, we discuss a number of improvements in time and space efficiency with regard to practical implementations.

Our algorithm, used in conjunction with higher-dimensional heuristics, is able to calculate for the first time the optimal alignment for almost all of the problems in Reference 1 of the benchmark database *BAliBASE*.

## 1. Introduction: Multiple Sequence Alignment

The *multiple sequence alignment problem (MSA)* in computational biology consists in aligning several sequences, e.g. related genes from different organisms, in order to reveal simi-larities and differences across the group. Either DNA can be directly compared, and the underlying alphabet $\Sigma$ consists of the set $\{C,G,A,T\}$ for the four standard nucleotide bases cytosine, guanine, adenine and thymine; or we can compare proteins, in which case $\Sigma$ comprises the twenty amino acids.

Roughly speaking, we try to write the sequences one above the other such that the columns with matching letters are maximized; thereby gaps (denoted here by an additional letter "_") may be inserted into either of them in order to shift the remaining characters into better corresponding positions. Different letters in the same column can be interpreted as being caused by point mutations during the course of evolution that substituted one amino acid by another one; gaps can be seen as insertions or deletions (since the direction of change is often not known, they are also collectively referred to as *indels*). Presumably, the alignment with the fewest mismatches or indels constitutes the biologically most plausible explanation.

There is a host of applications of MSA within computational biology; e.g., for determining the evolutionary relationship between species, for detecting functionally active sites which tend to be preserved best across homologous sequences, and for predicting three-dimensional protein structure.

Formally, one associates a cost with an alignment and tries to find the (mathematically) *optimal* alignment, i.e., that one with minimum cost. When designing a cost function, computational efficiency and biological meaning have to be taken into account. The most widely-used definition is the *sum-of-pairs* cost function. First, we are given a symmetric $(|\Sigma| + 1)^2$ matrix containing penalties (scores) for substituting a letter with another one (or a gap). In the simplest case, this could be one for a mismatch and zero for a match, but more biologically relevant scores have been developed. Dayhoff, Schwartz, and Orcutt (1978) have proposed a model of molecular evolution where they estimate the exchange probabilities of amino acids for different amounts of evolutionary divergence; this gives rise to the so-called *PAM matrices*, where PAM250 is generally the most widely used; Jones, Taylor, and Thornton (1992) refined the statistics based on a larger body of experimental data. Based on such a substitution matrix, the sum-of-pairs cost of an alignment is defined as the sum of penalties between all letter pairs in corresponding column positions.

A pairwise alignment can be conveniently depicted as a path between two opposite corners in a two-dimensional grid (Needleman and Wunsch, 1981): one sequence is placed on the horizontal axis from left to right, the other one on the vertical axis from top to bottom. If there is no gap in either string, the path moves diagonally down and right; a gap in the vertical (horizontal) string is represented as a horizontal (vertical) move right (down), since a letter is consumed in only one of the strings. The alignment graph is directed and acyclic, where a (non-border) vertex has incoming edges from the left, top, and top-left adjacent vertices, and outgoing edges to the right, bottom, and bottom-right vertices.

Pairwise alignment can be readily generalized to the simultaneous alignment of multiple sequences, by considering higher-dimensional lattices. For example, an alignment of three sequences can be visualized as a path in a cube. Fig. 1 illustrates an example for the strings `ABCB`, `BCD`, and `DB`. It also shows the computation of the sum-of-pairs cost, for a hypothetical substitution matrix. A real example (problem `2trx` of *BAliBASE*, see Sec. 7.3) is given in Fig. 2.

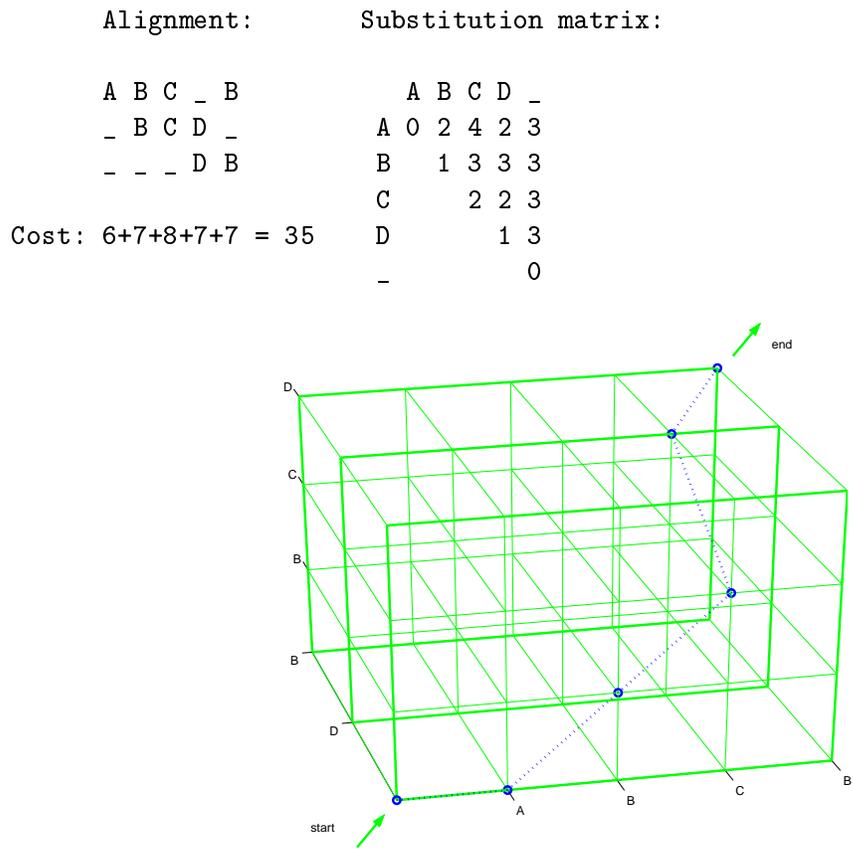

Figure 1: Fictitious alignment problem: Column representation, cost matrix, three-dimensional visualization of the alignment path through the cube.

A number of improvements can be integrated into the sum-of-pairs cost, like associating weights with sequences, and using different substitution matrices for sequences of varying evolutionary distance. A major issue in multiple sequence alignment algorithms is their ability to handle gaps. Gap penalties can be made dependent on the neighbor letters. Moreover, it has been found (Altschul, 1989) that assigning a fixed score for each indel sometimes does not produce the biologically most plausible alignment. Since the insertion of a sequence of $x$ letters is more likely than $x$ separate insertions of a single letter, gap cost functions have been introduced that depend on the length of a gap. A useful approximation are *affine gap costs*, which distinguish between opening and extension of a gap and charge $a + b * x$ for a gap of length $x$, for appropriate $a$ and $b$. Another frequently used modification is to waive the penalties for gaps at the beginning or end of a sequence.

Technically, in order to deal with affine gap costs we can no longer identify nodes in the search graph with lattice vertices, since the cost associated with an edge depends on the preceding edge in the path. Therefore, it is more suitable to store lattice edges in the priority

```
1thx    _aeqpvlvyfwaswcgpcqlmsplinlaantysdrlkvvkleidpnpttvkkyk______vegvpal
1grx    __mqtvi__fgrsgcpysvrakdlaeklsnerdd_fqyqyvdiraegitkedlqqkagkpvetvp__
1erv    agdklvvvdfsatwcgpckmikpffhslsekysn_viflevdvddcqdvasece______vksmptf
2trcP   _kvttivvniyedgvrgcdalnssleclaaeypm_vkfckira_sntgagdrfs______sdvlptl
```

```
1thx    rlvkgeqildstegvis__kdkllsf_ldthln_________
1grx    qifvdqqhiggytdfaawvken_____lda___________
1erv    qffkkgqkvgefsgan___kek_____leatine__lv____
2trcP   lvykggelisnfisvaeqfaedffaadvesflneygllper_
```

Figure 2: Alignment of problem `2trx` of *BAliBASE*, computed with algorithm settings as described in Sec. 7.3.

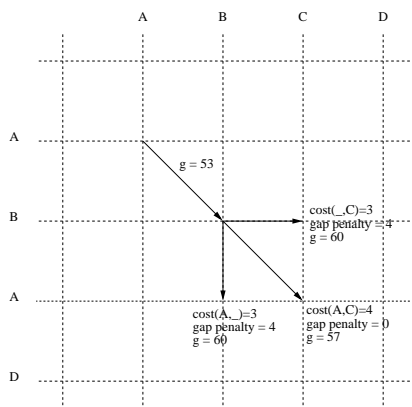

Figure 3: Example of computing path costs with affine gap function; the substitution matrix of Fig. 1 and a gap opening penalty of 4 is used.

queue, and let the transition costs for $u \rightarrow v, v \rightarrow w$ be the sum-of-pairs substitution costs for using one character from each sequence or a gap, plus the incurred gap penalties for $v \rightarrow w$ followed by $u \rightarrow v$. This representation was adopted in the program *MSA* (Gupta, Kececioglu, & Schaeffer, 1995). Note that the state space in this representation grows by a factor of $2^k$. An example of how successor costs are calculated, with the cost matrix of Fig. 1 and a gap opening penalty of 4, is shown in Fig. 3.

For convenience of terminology in the sequel we will still refer to nodes when dealing with the search algorithm.

## 2. Overview

Wang and Jiang (1994) have shown that the optimal multiple sequence alignment problem is *NP*-hard; therefore, we cannot hope to achieve an efficient algorithm for an arbitrary number of sequences. As a consequence, alignment tools most widely used in practice sacrifice the sound theoretical basis of exact algorithms, and are heuristic in nature (Chan, Wong, & Chiu, 1992). A wide variety of techniques has been developed. *Progressive* methods build up the alignment gradually, starting with the closest sequences and successively adding more distant ones. *Iterative* strategies refine an initial alignment through a sequence of improvement steps.

Despite their limitation to moderate number of sequences, however, the research into exact algorithms is still going on, trying to push the practical boundaries further. They still form the building block of heuristic techniques, and incorporating them into existing tools could improve them. For example, an algorithm iteratively aligning two groups of sequences at a time could do this with three or more, to better avoid local minima. Moreover, it is theoretically important to have the "gold standard" available for evaluation and comparison, even if not for all problems.

Since MSA can be cast as a minimum-cost path finding problem, it turns out that it is amenable to heuristic search algorithms developed in the AI community; these are actually among the currently best approaches. Therefore, while many researchers in this area have often used puzzles and games in the past to study heuristic search algorithms, recently there has been a rising interest in MSA as a testbed with practical relevance, e.g., (Korf, 1999; Korf & Zhang, 2000; Yoshizumi, Miura, & Ishida, 2000; Zhou & Hansen, 2003b); its study has also led to major improvements of general search techniques.

It should be pointed out that the definition of the MSA problem as given above is not the only one; it competes with other attempts at formalizing biological meaning, which is often imprecise or depends on the type of question the biologist investigator is pursuing. E.g., in this paper we are only concerned with *global alignment* methods, which find an alignment of entire sequences. Local methods, in contrast, are geared towards finding maximally similar partial sequences, possibly ignoring the remainder.

In the next section, we briefly review previous approaches, based on dynamic programming and incorporating lower and upper bounds. In Sec. 4, we describe a new algorithm that combines and extends some of these ideas, and allows to reduce the storage of *Closed* nodes by partially recomputing the solution path at the end (Sec. 5). Moreover, it turns out that our algorithm's iterative deepening strategy can be transferred to find a good balance between the computation of *improved heuristics* and the main search (Sec. 6), an issue that has previously been a major obstacle for their practical application. Sec. 7 presents an experimental comparison with *Partial Expansion A\** (Yoshizumi, Miura, & Ishida, 2000), one of the currently most successful approaches. We also solve all but two problems of Reference 1 of the widely used benchmark database *BAliBASE* (Thompson, Plewniak, & Poch, 1999). To the best of our knowledge, this has not been achieved previously with an exact algorithm.

## 3. Previous Work

A number of exact algorithms have been developed previously that can compute alignments of a moderate number of sequences. Some of them are mostly constrained by available memory, some by the required computation time, and some on both. We can roughly group them into two categories: those based on the dynamic programming paradigm, which proceed primarily in breadth-first fashion; and best-first search, utilizing lower and upper bounds to prune the search space. Some recent research, including our new algorithm introduced in Sec. 4, attempts to beneficially combine these approaches.

### 3.1 Dijkstra's Algorithm and Dynamic Programming

Dijkstra (1959) presented a general algorithm for finding the shortest (resp. minimum cost) path in a directed graph. It uses a *priority queue (heap)* to store nodes $v$ together with the shortest found distance from the start node $s$ (i.e., the top-left corner of the grid) to $v$ (also called the $g$-value of $v$). Starting with only $s$ in the priority queue, in each step, an edge with the minimum $g$-value is removed from the priority queue; its *expansion* consists in generating all of its successors (vertices to the right and/or below) reachable in one step, computing their respective $g$-value by adding the edge cost to the previous $g$-value, and inserting them in turn into the priority queue in case this newly found distance is smaller than their previous $g$-value. By the time a node is expanded, the $g$-value is guaranteed to be the minimal path cost from the start node, $g^*(v) = d(s, v)$. The procedure runs until the priority queue becomes empty, or the target node $t$ (the bottom-right corner of the grid) has been reached; its $g$-value then constitutes the optimal solution cost $g^*(t) = d(s, t)$ of the alignment problem. In order to trace back the path corresponding to this cost, we move backwards to the start node choosing predecessors with minimum cost. The nodes can either be stored in a fixed matrix structure corresponding to the grid, or they can be dynamically generated; in the latter case, we can explicitly store at each node a backtrack-pointer to this optimal parent.

For integer edge costs, the priority queue can be implemented as a bucket array pointing to doubly linked lists (Dial, 1969), so that all operations can be performed in constant time (To be precise, the *DeleteMin*-operation also needs a pointer that runs through all different $g$-values once; however, we can neglect this in comparison to the number of expansions). To expand a vertex, at most $2^k - 1$ successor vertices have to be generated, since we have the choice of introducing a gap in each sequence. Thus, Dijkstra's algorithm can solve the multiple sequence alignment problem in $O(2^k N^k)$ time and $O(N^k)$ space for $k$ sequences of length $\leq N$.

A means to reduce the number of nodes that have to be stored for path reconstruction is by associating a counter with each node that maintains the number of children whose backtrack-pointer refers to them (Gupta et al., 1995). Since each node can be expanded at most once, after this the number of referring backtrack-pointers can only decrease, namely, whenever a cheaper path to one of its children is found. If a node's reference count goes to zero, whether immediately after its expansion or when it later loses a child, it can be deleted for good. This way, we only keep nodes in memory that have at least one descendant currently in the priority queue. Moreover, auxiliary data structures for vertices and coordinates are most efficiently stored in tries (prefix trees); they can be equipped with reference counters as well and be freed accordingly when no longer used by any edge.

The same complexity as for Dijkstra's algorithm holds for *dynamic programming (DP)*; it differs from the former one in that it scans the nodes in a fixed order that is known beforehand (hence, contrary to the name the exploration scheme is actually static). The exact order of the scan can vary (e.g., row-wise or column-wise), as long as it is compatible with the topological ordering of the graph (e.g., for two sequences that the cells left, top, and diagonally top-left have been explored prior to a cell). One particular such ordering is that of *antidiagonals*, diagonals running from upper-right to lower-left. The calculation of the antidiagonal of a node merely amounts to summing up its $k$ coordinates.

Hirschberg (1975) noticed that in order to determine only the cost of the optimal alignment $g^*(t)$, it would not be necessary to store the whole matrix; instead, when proceeding e.g. by rows it suffices to keep track of only $k$ of them at a time, deleting each row as soon as the next one is completed. This reduces the space requirement by one dimension from $O(N^k)$ to $O(kN^{k-1})$. In order to recover the solution path at the end, re-computation of the lost cell values is needed. A *Divide-and-conquer*-strategy applies the algorithm twice to half the grid each, once in forward and once in backward direction, meeting at a fixed middle row. By adding the corresponding forward and backward distances in this middle row and finding the minimum, one cell lying on an optimal path can be recovered. This cell essentially splits the problem into two smaller subproblems, one from the upper left corner to it, and the other one to the lower right corner; they can be recursively solved using the same method. In two dimensions, the computation time is at most doubled, and the overhead reduces even more in higher dimensions.

The *FastLSA* algorithm (Davidson, 2001) further refines Hirschberg's algorithm by exploiting additionally available memory to store more than one node on an optimal path, thereby reducing the number of re-computations.

## 3.2 Algorithms Utilizing Bounds

While Dijkstra's algorithm and dynamic programming can be viewed as variants of *breadth-first search*, we achieve *best first search* if we expand nodes $v$ in the order of an estimate (lower bound) of the total cost of a path from $s$ to the $t$ passing through $v$. Rather than using the $g$-value as in Dijkstra's algorithm, we use $f(v) := g(v) + h(v)$ as the heap key, where $h(v)$ is a lower bound on the cost of an optimal path from $v$ to $t$. If $h$ is indeed *admissible*, then the first solution found is guaranteed to be optimal (Hart, Nilsson, & Raphael, 1968). This is the classical best-first search algorithm, the *A* *algorithm*, well known in the artificial intelligence community. In this context, the priority queue maintaining the generated nodes is often also called the *Open* list, while the nodes that have already been expanded and removed from it constitute the *Closed* list. Fig. 4 schematically depicts a snapshot during a two-dimensional alignment problem, where all nodes with $f$-value no larger than the current $f_{\min}$ have been expanded. Since the accuracy of the heuristic decreases with the distance to the goal, the typical 'onion-shaped' distribution results, with the bulk being located closer to the start node, and tapering out towards higher levels.

The $A^*$ algorithm can significantly reduce the total number of expanded and generated nodes; therefore, in higher dimensions it is clearly superior to dynamic programming. How-ever, in contrast to the Hirschberg algorithm, it still stores all of the explored nodes in the *Closed* list. Apart from keeping track of the solution path, this is necessary to prevent the search from "leaking back", in the following sense.

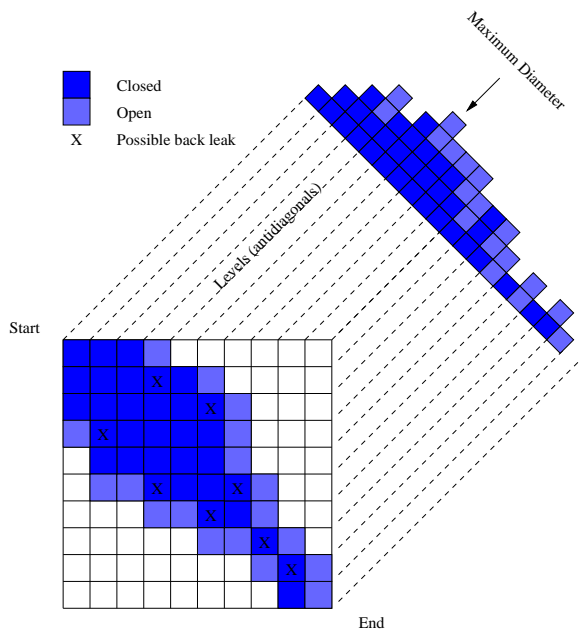

Figure 4: Snapshot during best-first search in pairwise alignment (schematically).

A heuristic $h$ is called *consistent* if $h(x) \leq h(x') + c(x, x')$, for any node $x$ and its child $x'$. A consistent heuristic ensures that (as in the case of Dijkstra's algorithm) at the time a node is expanded, its $g$-value is optimal, and hence it is never expanded again. However, if we try to delete the *Closed* nodes, then there can be topologically smaller nodes in *Open* with a higher $f$-value; when those are expanded at a later stage, they can lead to the re-generation of the node at a non-optimal $g$-value, since the first instantiation is no longer available for duplicate checking. In Fig. 4, nodes that might be subject to spurious re-expansion are marked "X".

Researchers have tried to avoid these leaks, while retaining the basic $A^*$ search scheme. Korf proposed to store a list of forbidden operators with each node, or to place the parents of a deleted node on *Open* with $f$-value infinity (Korf, 1999; Korf & Zhang, 2000). However, as Zhou and Hansen (2003a) remark, it is hard to combine this algorithm with techniques for reduction of the *Open* list, and moreover the storage of operators lets the size of the nodes grow exponentially with the number of sequences. In their algorithm, they keep track of the *kernel* of the *Closed* list, which is defined as the set of nodes that have only *Closed* nodes as parents; otherwise a *Closed* node is said to be in the *boundary*. The key idea is that only the boundary nodes have to be maintained, since they shield the kernel from re-expansions. Only when the algorithm gets close to the memory limit nodes from the kernel are deleted; the backtrack pointer of the children is changed to the parents of the deleted nodes, which become *relay nodes* for them. For the final reconstruction of the optimal solution path, the algorithm is called recursively for each relay node to bridge the gap of missing edges.

In addition to the *Closed* list, also the *Open* list can grow rapidly in sequence alignment problems. Particularly, since in the original $A^*$ algorithm the expansion of a node generates all of its children at once, those whose $f$-value is larger than the optimal cost $g^*(t)$ are kept in the heap up to the end, and waste much of the available space.

If an upper bound $U$ on the optimal solution cost $g^*(t)$ is known, then nodes $v$ with $f(v) > U$ can be pruned right away; this idea is used in several articles (Spouge, 1989; Gupta et al., 1995). One of the most successful approaches is Yoshizumi et al.'s (2000) *Partial Expansion $A^*$ ($PEA^*$)*. Each node stores an additional value $F$, which is the minimum $f$-value of all of its yet ungenerated children. In each step, only a node with minimum $F$-value is expanded, and only those children with $f = F$ are generated. This algorithm clearly only generates nodes with $f$ value no larger than the optimal cost, which cannot be avoided altogether. However, the overhead in computation time is considerable: in the straightforward implementation, if we want to maintain nodes of constant size, generating one edge requires determining the $f$-values of all successors, such that for an interior node which eventually will be fully expanded the computation time is of the order of the square of the number of successors, which grows as $O(2^k)$ with the number of sequences $k$. As a remedy, in the paper it is proposed to relax the condition by generating all children with $f \leq F + C$, for some small $C$.

An alternative general search strategy to $A^*$ that uses only linear space is *iterative deepening $A^*$ ($IDA^*$)* (Korf, 1985). The basic algorithm conducts a depth-first search up to a pre-determined threshold for the $f$-value. During the search, it keeps track of the smallest $f$-value of a generated successor that is larger than the threshold. If no solution is found, this provides an increased threshold to be used in the next search iteration.

Wah and Shang (1995) suggested more liberal schemes for determining the next threshold dynamically in order to minimize the number of recomputations. $IDA^*$ is most efficient in tree structured search spaces. However, it is difficult to detect duplicate expansions without additional memory; Therefore, unfortunately it is not applicable in lattice-structured graphs like in the sequence alignment problem due to the combinatorially explosive number of paths between any two given nodes.

A different line of research tries to restrict the search space of the breadth-first approaches by incorporating bounds. Ukkonen (1985) presented an algorithm for the pairwise alignment problem which is particularly efficient for similar sequences; its computation time scales as $O(dm)$, where $d$ is the optimal solution cost. First consider the problem of deciding whether a solution exists whose cost is less than some upper threshold $U$. We can restrict the evaluation of the $DP$ matrix to a band of diagonals where the minimum number of indels required to reach the diagonal, times the minimum indel cost, does not exceed $U$. In general, starting with a minimum $U$ value, we can successively double $G$ until the test returns a solution; the increase of computation time due to the recomputations is then also bounded by a factor of 2.

Another approach for multiple sequence alignment is to make use of the lower bounds $h$ from $A^*$. The key idea is the following: Since all nodes with an $f$-value lower than $g^*(t)$ have to be expanded anyway in order to guarantee optimality, we might as well explore them in

595

any reasonable order, like that of Dijkstra's algorithm or *DP*, if we only knew the optimal cost. Even slightly higher upper bounds will still help pruning. Spouge (1989) proposed to bound *DP* to vertices $v$ where $g(v) + h(v)$ is smaller than an upper bound for $g^*(t)$.

*Linear Bounded Diagonal Alignment (LBD-Align)* (Davidson, 2001) uses an upper bound in order to reduce the computation time and memory in solving a pairwise alignment problem by dynamic programming. The algorithm calculates the *DP* matrix one antidiagonal at a time, starting in the top left corner, and working down towards bottom-right. While $A^*$ would have to check the bound in every expansion, *LBD-Align* only checks the top and bottom cell of each diagonal. If e.g. the top cell of a diagonal has been pruned, all the remaining cells in that row can be pruned as well, since they are only reachable through it; this means that the pruning frontier on the next row can be shifted down by one. Thus, the pruning overhead can be reduced from a quadratic to a linear amount in terms of the sequence length.

### 3.3 Obtaining Heuristic Bounds

Up to now we have assumed lower and upper bounds, without specifying how to derive them. Obtaining an inaccurate *upper bound* on $g^*(t)$ is fairly easy, since we can use the cost of *any* valid path through the lattice. Better estimates are e.g. available from heuristic linear-time alignment programs such as *FASTA* and *BLAST* (Altschul, Gish, Miller, Myers, & Lipman, 1990), which are a standard method for database searches. Davidson (2001) employed a local beam search scheme.

Gusfield (1993) proposed an approximation called the *star-alignment*. Out of all the sequences to be aligned, one *consensus sequence* is chosen such that the sum of its pairwise alignment costs to the rest of the sequences is minimal. Using this "best" sequence as the center, the other ones are aligned using the "once a gap, always a gap" rule. Gusfield showed that the cost of the optimal alignment is greater or equal to the cost of this star alignment, divided by $(2 - 2/k)$.

For use in heuristic estimates, lower bounds on the $k$-alignment are often based on optimal alignments of subsets of $m < k$ sequences. In general, for a vertex $v$ in $k$-space, we are looking for a lower bound for a path from $v$ to the target corner $t$. Consider first the case $m = 2$. The cost of such a path is, by definition, the sum of its edge costs, where each edge cost in turn is the sum of all pairwise (replacement or gap) penalties. Each multiple sequence alignment induces a pairwise alignment for sequences $i$ and $j$, by simply copying rows $i$ and $j$ and ignoring columns with a "_" in both rows. These pairwise alignments can be visualized as the projection of an alignment onto its *faces*, cf. Fig. 1.

By interchange of the summation order, the sum-of-pairs cost is the sum of all pairwise alignment costs of the respective paths projected on a face, each of which cannot be smaller than the optimal pairwise path cost. Thus, we can construct an admissible heuristic $h_{pair}$ by computing, for each pairwise alignment and for each cell in a pairwise problem, the cheapest path cost to the goal node.

The optimal solutions to all pairwise alignment problems needed for the lower bound $h$ values are usually computed prior to the main search in a preprocessing step (Ikeda & Imai, 1994). To this end, it suffices to apply the ordinary *DP* procedure; however, since this time we are interested in the lowest cost of a path from $v$ to $t$, it runs in backward direction, proceeding from the lower right corner to the upper left, expanding all possible parents of a vertex in each step.

Let $U$ be an upper bound on the cost of an optimal multiple sequence alignment $G$. The sum of all optimal alignment costs $L_{ij} = d(s_{ij}, t_{ij})$ for pairwise subproblems $i, j \in \{1, \ldots, k\}, i < j$, call it $L$, is a lower bound on $G$. Carrillo and Lipman (1988) pointed out that by the additivity of the sum-of-pairs cost function, any pairwise alignment induced by the optimal multiple sequence alignment can at most be $\delta = U - L$ larger than the respective optimal pairwise alignment. This bound can be used to restrict the number of values that have to be computed in the preprocessing stage and have to be stored for the calculation of the heuristic: for the pair of sequences $i, j$, only those nodes $v$ are feasible such that a path from the start node $s_{ij}$ to the goal node $t_{ij}$ exists with total cost no more than $L_{i,j} + \delta$. To optimize the storage requirements, we can combine the results of two searches. First, a forward pass determines for each relevant node $v$ the minimum distance $d(s_{ij}, v)$ from the start node. The subsequent backward pass uses this distance like an 'exact heuristic' and stores the distance $d(v, t_{ij})$ from the target node only for those nodes with $d(s_{ij}, v) + d(v, t_{ij}) \leq d(s, t) + \delta$[1].

Still, for larger alignment problems the required storage size can be extensive. The program *MSA* (Gupta et al., 1995) allows the user to adjust $\delta$ to values below the Carrillo-Lipman bound individually for each pair of sequences. This makes it possible to generate at least heuristic alignments if time or memory doesn't allow for the complete solution; moreover, it can be recorded during the search if the $\delta$-bound was actually reached. In the negative case, optimality of the found solution is still guaranteed; otherwise, the user can try to run the program again with slightly increased bounds.

The general idea of precomputing simplified problems and storing the solutions for use as a heuristic has been explored under the name of *pattern databases* (Culberson & Schaeffer, 1998). However, these approaches implicitly assume that the computational cost can be amortized over many search instances to the same target. In contrast, in the case of MSA, the heuristics are instance-specific, so that we have to strike a balance. We will discuss this in greater depth in Sec. 6.2.

## 4. Iterative-Deepening Dynamic Programming

As we have seen, a fixed search order as in dynamic programming can have several advantages over pure best-first selection.

- Since *Closed* nodes can never be reached more than once during the search, it is safe to delete useless ones (those that are not part of any shortest path to the current *Open* nodes) and to apply path compression schemes, such as the Hirschberg algorithm. No sophisticated schemes for avoiding 'back leaks' are required, such as the above-mentioned methods of core set maintenance and dummy node insertion into *Open*.

---

1. A slight technical complication arises for affine gap costs: recall that *DP* implementations usually charge the gap opening penalty to the $g$-value of the edge $e$ starting the gap, while the edge $e'$ ending the gap carries no extra penalty at all. However, since the sum of pairs heuristics $h$ is computed in backward direction, using the same algorithm we would assign the penalty for the same path instead to $e'$. This means that the heuristic $f = g + h$ would no longer be guaranteed to be a lower bound, since it contains the penalty twice. As a remedy, it is necessary to make the computation symmetric by charging both the beginning and end of a gap with half the cost each. The case of the beginning and end of the sequences can be handled most conveniently by starting the search from a "dummy" diagonal edge $((-1, \ldots, -1), (0, \ldots, 0))$, and defining the target edge to be the dummy diagonal edge $((N, \ldots, N), (N+1, \ldots, N+1))$, similar to the arrows shown in Fig. 1.

- Besides the size of the *Closed* list, the memory requirement of the *Open* list is determined by the *maximum* number of nodes that are open *simultaneously at any time* while the algorithm is running. When the $f$-value is used as the key for the priority queue, the *Open* list usually contains all nodes with $f$-values in some range $(f_{min}, f_{min} + \delta)$; this set of nodes is generally spread across all over the search space, since $g$ (and accordingly $h = (f - g)$) can vary arbitrarily between 0 and $f_{min} + \delta$. As opposed to that, if $DP$ proceeds along levels of antidiagonals or rows, at any iteration at most $k$ levels have to be maintained at the same time, and hence the size of the *Open* list can be controlled more effectively. In Fig. 4, the pairwise alignment is partitioned into antidiagonals: the maximum number of open nodes in any two adjacent levels is four, while the total amounts to seventeen[2].

- For practical purposes, the running time should not only be measured in terms of the number of node expansions, but one should also take into account the *execution time* needed for an expansion. By arranging the exploration order such that edges with the same head node (or more generally, those sharing a common coordinate prefix) are dealt with one after the other, much of the computation can be cached, and edge generation can be sped up significantly. We will come back to this point in Sec. 6.

The remaining issue of a static exploration scheme consists in adequately bounding the search space using the $h$-values. $A^*$ is known to be minimal in terms of the number of node expansions. If we knew the cost $g^*(t)$ of a cheapest solution path beforehand, we could simply proceed level by level of the grid, however only immediately prune generated edges $e$ whenever $f(e) > g^*(t)$. This would ensure that we only generate those edges that would have been generated by algorithm $A^*$, as well. An upper threshold would additionally help reduce the size of the *Closed* list, since a node can be pruned if all of its children lie beyond the threshold; additionally, if this node is the only child of its parent, this can give rise to a propagating chain of ancestor deletions.

We propose to apply a search scheme that carries out a series of searches with successively larger thresholds, until a solution is found (or we run out of memory or patience). The use of such an upper bound parallels that in the $IDA^*$ algorithm.

The resulting algorithm, which we will refer to as *Iterative-Deepening Dynamic Programming (IDDP)*, is sketched in Fig. 5. The outer loop initializes the threshold with a lower bound (e.g., $h(s)$), and, unless a solution is found, increases it up to an upper bound. In the same manner as in the $IDA^*$ algorithm, in order to make sure that at least one additional edge is explored in each iteration the threshold has to be increased correspondingly at least to the minimum cost of a fringe edge that exceeded the previous threshold. This fringe increment is maintained in the variable *minNextThresh*, initially estimated as the upper bound, and repeatedly decreased in the course of the following expansions.

---

2. Contrary to what the figure might suggest, $A^*$ can open more than two nodes per level in pairwise alignments, if the set of nodes no worse than some $f_{\min}$ contains "holes".

```
procedure IDDP(Edge startEdge, Edge targetEdge, int lowerBound, int upperBound)
int thresh = lowerBound
{Outer loop: Iterative deepening phases}
while (thresh ≤ upperBound) do
  Heap h = {(startEdge, 0)}
  int minNextThresh = upperBound
  {Inner loop: Bounded dynamic programming}
  while (not h.IsEmpty()) do
    Edge e = h.DeleteMin() {Find and remove an edge with minimum level}
    if (e == targetEdge) then
       {Optimal alignment found}
       return TraceBackPath(startEdge, targetEdge)
    end if
    Expand(e, thresh, minNextThresh)
  end while
  int threshIncr = ComputeThreshIncr() {Compute search threshold for next iteration, see text}
  thresh = max(thresh + threshIncr, minNextThresh)
end while
print("No alignment with cost at most upperBound found")
```

Figure 5: Algorithm *Iterative-Deepening Dynamic Programming*.

In each step of the inner loop, we select and remove a node from the priority queue whose level is minimal. As explained later in Sec. 6, it is favorable to break ties according to the lexicographic order of target nodes. Since the total number of possible levels is comparatively small and known in advance, the priority queue can be implemented using an array of linked lists (Dial, 1969); this provides constant time operations for insertion and deletion.

The expansion of an edge $e$ is partial (Fig. 6). A child edge might already exist from an earlier expansion of an edge with the same head vertex; we have to test if we can decrease the $g$-value. Otherwise, we generate a new edge, if only temporarily for the sake of calculating its $f$-value; that is, if its $f$-value exceeds the search threshold of the current iteration, its memory is immediately reclaimed. Moreover, in this case the fringe threshold *minNext-Thresh* is updated. In a practical implementation, we can prune unnecessary accesses to partial alignments *inside* the calculation of the heuristic $e.GetH()$ as soon as as the search threshold has already been reached.

The relaxation of a child edge within the threshold is performed by the subprocedure *UpdateEdge* (cf. Fig. 7). This is similar to the corresponding relaxation step in $A^*$, updating the child's $g$- and $f$ values, its parent pointers, and inserting it into *Open*, if not already contained. However, in contrast to best-first search, it is inserted into the heap according to the antidiagonal level of its head vertex. Note that in the event that the former parent loses its last child, propagation of deletions (Fig. 8) can ensure that only those *Closed* nodes continue to be stored that belong to some solution path. Edge deletions can also ensue deletion of dependent vertex and coordinate data structures (not shown in the pseudocode). The other situation that gives rise to deletions is if immediately after the expansion of a node no children are pointing back to it (the children might either be reachable more cheaply from different nodes, or their $f$-value might exceed the threshold).

**procedure** Expand(Edge e, int thresh, int minNextThresh)
**for all** Edge child ∈ Succ(e) **do**
   {Retrieve child or tentatively generate it if not yet existing, set boolean variable 'created' accordingly}
   int newG = e.GetG() + GapCost(e, child)
           + child.GetCost()
   int newF = newG + child.GetH()
   **if** (newF ≤ thresh **and** newG < child.GetG()) **then**
     {Shorter path than current best found, estimate within threshold}
     child.SetG(newG)
     UpdateEdge(e, child, h) {Update search structures}
   **else if** (newF > thresh) **then**
     minNextThresh =
       min(minNextThresh, newF)
     {Record minimum of pruned edges}
     **if** (created) **then**
       Delete(child) {Make sure only promising edges are stored}
     **end if**
   **end if**
**end for**
**if** (e.ref == 0) **then**
   DeleteRec(e) {No promising children could be inserted into the heap}
**end if**

Figure 6: Edge expansion in *IDDP*.

**procedure** UpdateEdge(Edge parent, Edge child, Heap h)
parent.ref++
child.GetBacktrack().ref−−
**if** (child.GetBacktrack().ref == 0) **then**
   DeleteRec(child.GetBacktrack()) {The former parent has lost its last child and becomes useless}
**end if**
child.SetBacktrack(parent)
**if** (**not** h.Contains(child)) **then**
   h.Insert(child, child.GetHead().GetLevel())
**end if**

Figure 7: Edge relaxation in *IDDP*.

The correctness of the algorithm can be shown analogously to the soundness proof of $A^*$. If the threshold is smaller than $g^*(t)$, the *DP* search will terminate without encountering a solution; otherwise, only nodes are pruned that cannot be part of an optimal path. The invariant holds that there is always a node in each level which lies on an optimal path and is in the *Open* list. Therefore, if the algorithm terminates only when the heap runs empty, the best found solution will indeed be optimal.

The iterative deepening strategy results in an overhead computation time due to re-expansions, and we are trying to restrict this overhead as much as possible. More precisely, **procedure** DeleteRec(Edge e)
**if** (e.GetBacktrack() $\neq$ **nil**) **then**
   e.GetBacktrack().ref$--$
   **if** (e.GetBacktrack().ref $== 0$) **then**
      DeleteRec(e.GetBacktrack())
   **end if**
**end if**
Delete(e)

Figure 8: Recursive deletion of edges that are no longer part of any solution path.

**procedure** TraceBack(Edge startEdge, Edge e)
**if** (e == startEdge) **then**
   **return** {End of recursion}
**end if**
**if** (e.GetBackTrack().GetTarget() $\neq$ e.GetSource()) **then**
   {Relay node: recursive path reconstruction}
   IDDP( e.GetBackTrack(), e, e.GetF(), e.GetF())
**end if**
OutputEdge(e)
TraceBack(startEdge, e.GetBackTrack())

Figure 9: Divide-and-Conquer solution reconstruction in reverse order.

we want to minimize the ratio

$$\nu = \frac{n_{\text{IDDP}}}{n_{A^*}},$$

where $n_{\text{IDDP}}$ and $n_{A^*}$ denote the number of expansions in *IDDP* and $A^*$, respectively. One way to do so (Wah & Shang, 1995) is to choose a threshold sequence $\theta_1, \theta_2, \dots$ such that the number of expansions $n_i$ in stage $i$ satisfies

$$n_i = r n_{i-1},$$

for some fixed ratio $r$. If we choose $r$ too small, the number of re-expansions and hence the computation time will grow rapidly, if we choose it too big, then the threshold of the last iteration can exceed the optimal solution cost significantly, and we will explore many irrelevant edges. Suppose that $n_0 r^p < n_{A^*} \leq n_0 r^{p+1}$. Then the algorithm performs $p + 1$ iterations. In the worst case, the overshoot will be maximal if $A^*$ finds the optimal solution just above the previous threshold, $n_{A^*} = n_0 r^p + 1$. The total number of expansions is $n_0 \sum_{i=0}^{p+1} r^i = n_0 \frac{r(r^{p+1}-1)}{r-1}$, and the ratio $\nu$ becomes approximately $\frac{r^2}{r-1}$. By setting the derivative of this expression to zero, we find that the optimal value for $r$ is 2; the number of expansions should double from one search stage to the next. If we achieve doubling, we will expand at most four times as many nodes as $A^*$.

Like in Wah and Shang's (1995) scheme, we dynamically adjust the threshold using runtime information. Procedure *Compute ThreshIncr* stores the sequence of expansion numbers and thresholds from the previous search stages, and then uses curve fitting for extrapolation (in the first few iterations without sufficient data available, a very small default threshold is applied). We found that the distribution of nodes $n(\theta)$ with $f$-value smaller or equal to threshold $\theta$ can be modeled very accurately according to the exponential approach

$$n(\theta) = A \cdot B^{\theta}.$$

Consequently, in order to attempt to double the number of expansions, we choose the next threshold according to

$$\theta_{i+1} = \theta_i + \frac{1}{\log_2 B}.$$

## 5. Sparse Representation of Solution Paths

When the search progresses along antidiagonals, we do not have to fear back leaks, and are free to prune *Closed* nodes. Similarly as in Zhou and Hansen's (2003a) work, however, we only want to delete them lazily and incrementally when being forced by the algorithm approaching the computer's memory limit.

When deleting an edge $e$, the backtrack-pointers of its child edges that refer to it are redirected to the respective predecessor of $e$, whose reference count is increased accordingly. In the resulting *sparse solution path* representation, backtrack pointers can point to any optimal ancestors.

After termination of the main search, we trace back the pointers starting with the goal edge; this is outlined in Procedure *TraceBack* (Fig. 9), which prints out the solution path in reverse order. Whenever an edge $e$ points back to an ancestor $e'$ which is not its direct parent, we apply an auxiliary search from start edge $e'$ to goal edge $e$ in order to reconstruct the missing links of the optimal solution path. The search threshold can now be fixed at the known solution cost; moreover, the auxiliary search can prune those edges that cannot be ancestors of $e$ because they have some coordinate greater than the corresponding coordinate in $e$. Since also the shortest distance between $e$ and $e'$ is known, we can stop at the first path that is found at this cost. To improve the efficiency of the auxiliary search even further, the heuristic could be recomputed to suit the new target. Therefore, the cost of restoring the solution path is usually marginal compared to that of the main search.

Which edges are we going to prune, in which order? For simplicity, assume for the moment that the *Closed* list consists of a single solution path. According to the Hirschberg approach, we would keep only one edge, preferably lying near the center of the search space (e.g., on the longest anti-diagonal), in order to minimize the complexity of the two auxiliary searches. With additional available space allowing to store three relay edges, we would divide the search space into four subspaces of about equal size (e.g., additionally storing the antidiagonals half-way between the middle antidiagonal and the start node resp. the target node). By extension, in order to incrementally save space under diminishing resources we would first keep only every other level, then every fourth, and so on, until only the start edge, the target edge, and one edge half-way on the path would be left.

Since in general the *Closed* list contains multiple solution paths (more precisely, a tree of solution paths), we would like to have about the same density of relay edges on each of them. For the case of $k$ sequences, an edge reaching level $l$ with its head node can originate with its tail node from level $l - 1, \dots, l - k$. Thus, not every solution path passes through each level, and deleting every other level could result in leaving one path completely intact, while extinguishing another totally. Thus, it is better to consider contiguous *bands* of $k$

```
procedure SparsifyClosed()
for (int sparse = 1 to ⌊log₂ N⌋) do
  while (UsedMemory() > maxMemory and exists {Edge e ∈ Open | e.GetLastSparse() <
  sparse}) do
    Edge pred = e.GetBacktrack()
    {Trace back solution path}
    while (pred ≠ nil and e.GetLastSparse() < sparse) do
      e.SetLastSparse(sparse) {Mark to avoid repeated trace-back}
      if (⌊pred.GetHead().GetLevel() / k⌋ mod 2^sparse ≠ 0) then
        {pred lies in prunable band: redirect pointer}
        e.SetBacktrack(pred.GetBacktrack())
        e.GetBacktrack().ref++
        pred.ref−−
        if (pred.ref == 0) then
          {e is the last remaining edge referring to pred}
          DeleteRec(pred)
        end if
      else
        {Not in prunable band: continue traversal}
        e = e.GetBacktrack()
      end if
      pred = e.GetBacktrack()
    end while
  end while
end for
```

Figure 10: Sparsification of *Closed* list under restricted memory.

levels each, instead of individual levels. Bands of this size cannot be skipped by any path. The total number of antidiagonals in an alignment problem of $k$ sequences of length $N$ is $k \cdot N - 1$; thus, we can decrease the density in $\lfloor \log_2 N \rfloor$ steps.

A technical implementation issue concerns the ability to enumerate all edges that reference some given prunable edge, without explicitly storing them in a list. However, the reference counting method described above ensures that any *Closed* edge can be reached by following a path bottom-up from some edge in *Open*. The procedure is sketched in Fig. 10. The variable *sparse* denotes the interval between level bands that are to be maintained in memory. In the inner loop, all paths to *Open* nodes are traversed in backward direction; for each edge $e'$ that falls into a prunable band, the pointer of the successor $e$ on the path is redirected to its respective backtrack pointer. If $e$ was the last edge referencing $e'$, the latter one is deleted, and the path traversal continues up to the start edge. When all *Open* nodes have been visited and the memory bound is still exceeded, the outer loop tries to double the number of prunable bands by increasing *sparse*.

Procedure *SparsifyClosed* is called regularly during the search, e.g., after each expansion. However, a naive version as described above would incur a huge overhead in computation time, particularly when the algorithm's memory consumption is close to the limit. Therefore, some optimizations are necessary. First, we avoid tracing back the same solution path at the same (or lower) *sparse* interval by recording for each edge the interval when it was traversed the last time (initially zero); only for an increased variable *sparse* there can be anything left for further pruning. In the worst case, each edge will be inspected $\lfloor \log_2 N \rfloor$ times. Secondly, it would be very inefficient to actually inspect each *Open* node in the inner loop, just to find that its solution path has been traversed previously, at the same or higher *sparse* value; however, with an appropriate bookkeeping strategy it is possible to reduce the time for this search overhead to $O(k)$.

## 6. Use of Improved Heuristics

As we have seen, the estimator $h_{pair}$, the sum of optimal pairwise goal distances, gives a lower bound on the actual path length. However, more powerful heuristics are also conceivable. While their computation will require more resources, the trade-off can prove itself worthwhile; the tighter the estimator is, the smaller is the space that the main search needs to explore.

### 6.1 Beyond Pairwise Alignments

Kobayashi and Imai (1998) suggested to generalize $h_{pair}$ by considering optimal solutions for subproblems of size $m > 2$. They proved that the following heuristics are admissible and more informed than the pairwise estimate.

- $h_{all,m}$ is the sum of all $m$-dimensional optimal costs, divided by $\binom{k-2}{m-2}$.

- $h_{one,m}$ splits the sequences into two sets of sizes $m$ and $k-m$; the heuristic is the sum of the optimal cost of the first subset, plus that of the second one, plus the sum of all 2-dimensional optimal costs of all pairs of sequences in different subsets. Usually, $m$ is chosen close to $k/2$.

These improved heuristics can reduce the main search effort by orders of magnitudes. However, in contrast to pairwise sub-alignments, time and space resources devoted to compute and store higher-dimensional heuristics are in general no longer negligible compared to the main search. Kobayashi and Imai (1998) noticed that even for the case $m = 3$ of triples of sequences, it can be impractical to compute the entire subheuristic $h_{all,m}$. As one reduction, they show that it suffices to restrict oneself to nodes where the path cost does not exceed the optimal path cost of the subproblem by more than

$$\delta = \binom{k-2}{m-2} U - \sum_{i_1,\ldots,i_m} d(s_{i_1,\ldots,i_m}, t_{i_1,\ldots,i_m});$$

this threshold can be seen as a generalization of the Carrillo-Lipman bound. However, it can still incur excessive overhead in space and computation time for the computation of the $\binom{k}{m}$ lower-dimensional subproblems. A drawback is that it requires an upper bound $U$, on whose accuracy also the algorithm's efficiency hinges. We could improve this bound by applying more sophisticated heuristic methods, but it seems counterintuitive to spend more time doing so which we would rather use to calculate the exact solution. In spite of its advantages for the main search, the expensiveness of the heuristic calculation appears as a major obstacle.

McNaughton, Lu, Schaeffer, and Szafron (2002) suggested to partition the heuristic into (hyper-) cubes using a hierarchical *oct-tree* data structure; in contrast to "full" cells, "empty" cells only retain the values at their surface. When the main search tries to use one of them, its interior values are recomputed on demand. Still, this work assumes that each node in the entire heuristic is calculated at least once using dynamic programming.

We see one cause of the dilemma in the implicit assumption that a *complete* computation is necessary. The bound $\delta$ above refers to the worst-case, and can generally include many more nodes than actually required in the main search. However, since we are only dealing with the heuristic, we can actually afford to miss some values occasionally; while this might slow down the main search, it cannot compromise the optimality of the final solution. Therefore, we propose to generate the heuristics with a much smaller bound $\delta$. Whenever the attempt to retrieve a value of the $m$-dimensional subheuristic fails during the main search, we simply revert to replacing it by the sum of the $\binom{m}{2}$ optimal pairwise goal distances it covers.

We believe that the *IDDP* algorithm lends itself well to make productive use of higher-dimensional heuristics. Firstly and most importantly, the strategy of searching to adaptively increasing thresholds can be transferred to the $\delta$-bound as well; this will be addressed in more detail in the next section.

Secondly, as far as a practical implementation is concerned, it is important to take into account not only how a higher-dimensional heuristic affects the number of node expansions, but also their time complexity. This time is dominated by the number of accesses to sub-alignments. With $k$ sequences, in the worst case an edge has $2^k - 1$ successors, leading to a total of

$$(2^k - 1)\binom{k}{m}$$

evaluations for $h_{all,m}$. One possible improvement is to enumerate all edges emerging from a given vertex in lexicographic order, and to store partial sums of heuristics of prefix subsets of sequences for later re-use. In this way, if we allow for a cache of linear size, the number of accesses is reduced to

$$\sum_{i=m}^{i=k} 2^i \binom{i-1}{m-1};$$

correspondingly, for a quadratic cache we only need

$$\sum_{i=m}^{i=k} 2^i \binom{i-2}{m-2}$$

evaluations. For instance, in aligning 12 sequences using $h_{all,3}$, a linear cache reduces the evaluations to about 37 percent within one expansion.

As mentioned above, in contrast to $A^*$, *IDDP* gives us the freedom to choose any particular expansion order of the edges within a given level. Therefore, when we sort edges lexicographically according to the target nodes, much of the cached prefix information can be shared additionally across consecutively expanded edges. The higher the dimension of the subalignments, the larger are the savings. In our experiments, we experienced speedups of up to eighty percent in the heuristic evaluation.

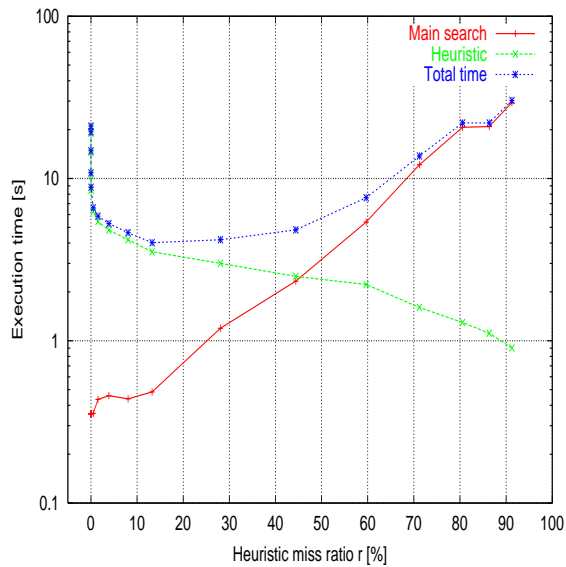

Figure 11: Trade-off between heuristic and main search: Execution times for problem `1tvxA` as a function of heuristic miss ratio.

## 6.2 Trade-Off between Computation of Heuristic and Main Search

As we have seen, we can control the size of the precomputed sub-alignments by choosing the bound $\delta$ up to which $f$-values of edges are generated beyond the respective optimal solution cost. There is obviously a trade-off between the auxiliary and main searches. It is instructive to consider the *heuristic miss ratio* $r$, i.e., the fraction of calculations of the heuristic $h$ during the main search when a requested entry in a partial MSA has not been precomputed. The optimum for the main search is achieved if the heuristic has been computed for every requested edge ($r = 0$). Going beyond that point will generate an unnecessarily large heuristic containing many entries that will never be actually used. On the other hand, we are free to allocate less effort to the heuristic, resulting in $r > 0$ and consequently decreasing performance of the main search. Generally, the dependence has an S-shaped form, as exemplified in Fig. 11 for the case of problem `1tvxA` of *BAliBASE* (cf. next section). Here, the execution time of one iteration of the main search at a fixed threshold of 45 above the lower bound is shown, which includes the optimal solution.

Fig. 11 illustrates the overall time trade-off between auxiliary and main search, if we fix $\delta$ at different levels. The minimum total execution time, which is the sum of auxiliary and main search, is attained at about $r = 0.15$ (5.86 seconds). The plot for the corresponding memory usage trade-off has a very similar shape.

Unfortunately, in general we do not know in advance the right amount of auxiliary search. As mentioned above, choosing $\delta$ according to the Carrillo-Lipman bound will ensure that every requested sub-alignment cost will have been precomputed; however, in general we will considerably overestimate the necessary size of the heuristic.

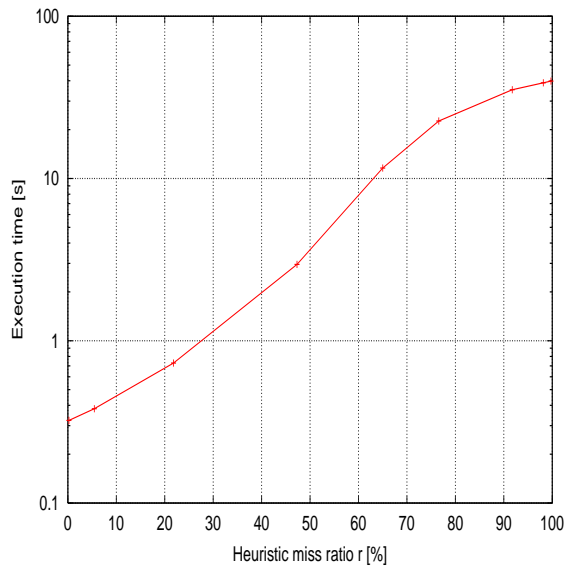

Figure 12: Time of the last iteration in the main search for problem `1tvxA` as a function of heuristic miss ratio.

As a remedy, our algorithm *IDDP* gives us the opportunity to recompute the heuristic in each threshold iteration in the main search. In this way, we can adaptively strike a balance between the two.

When the currently experienced miss rate $r$ rises above some threshold, we can suspend the current search, recompute the pairwise alignments with an increased threshold $\delta$, and resume the main search with the improved heuristics.

Like for the main search, we can accurately predict the auxiliary computation time and space at threshold $\delta$ using exponential fitting. Due to the lower dimensionality, it will generally increase less steeply; however, the constant factor might be higher for the heuristic, due to the combinatorial number of $\binom{k}{m}$ alignment problems to be solved.

A doubling scheme as explained above can bound the overhead to within a constant factor of the effort in the last iteration. In this way, when also limiting the heuristic computation time by a fixed fraction of the main search, we can ensure as an expected upper bound that the overall execution time stays within a constant factor of the search time that would be required using only the pairwise heuristic.

If we knew the exact relation between $\delta$, $r$, and the speedup of the main search, an ideal strategy would double the heuristic whenever the expected computation time is smaller than the time saved in the main search. However, as illustrated in Fig. 12, this dependence is more complex than simple exponential growth, it varies with the search depth and specifics of the problem. Either we would need a more elaborate model of the search space, or the algorithm would have to conduct exploratory searches in order to estimate the relation. We leave this issue to future work, and restrict ourselves here to a simplified, conservative heuristic: We hypothesize that the main search can be made twice as fast by a heuristic doubling if the miss rate $r$ rises above 25 percent; in our experiments, we found that this assumption is almost always true. In this event, since the effective branching factor of the main search is reduced by the improved heuristic, we also ignore the history of main search times in the exponential extrapolation procedure for subsequent iterations.

## 7. Experimental Results

In the following, we compare *IDDP* to one of the currently most successful approaches, *Partial Expansion A\**. We empirically explore the benefit of higher-dimensional heuristics; finally, we show its feasibility by means of the benchmark database *BAliBASE*.

### 7.1 Comparison to Partial Expansion A\*

For the first series of evaluations, we ran *IDDP* on the same set of sequences as chosen by Yoshizumi et al. (2000) (elongation factors EF-TU and EF-1$\alpha$ from various species, with a high degree of similarity). As in this work, substitution costs were chosen according to the PAM-250 matrix. The applied heuristic was the sum of optimal pairwise goal distances. The expansion numbers do not completely match with their results, however, since we applied the biologically more realistic affine gap costs: gaps of length $x$ were charged $8 + 8 \cdot x$, except at the beginning and end of a sequence, where the penalty was $8 \cdot x$.

All of the following experiments were run under RedHat Linux 7.3 on an Intel Xeon$^{TM}$ CPU with 3.06 GHz, and main memory of 2 Gigabytes; we used the gcc 2.96 compiler.

The total space consumption of a search algorithm is determined by the peak number of *Open* and *Closed* edges over the entire running time. Table 1 and Fig. 13 give these values for the series of successively larger sets of input sequences (with the sequences numbered as defined in Yoshizumi et al., 2000) $1 - 4$, $1 - 5$, ..., $1 - 12$.

With our implementation, the basic $A^*$ algorithm could be carried out only up to 9 sequences, before exhausting our computer's main memory.

Confirming the results of Yoshizumi et al. (2000), Partial Expansion requires only about one percent of this space. Interestingly, during the iteration with the peak in total numbers of nodes held in memory, no nodes are actually closed except in problem 6. This might be explained with the high degree of similarity between sequences in this example. Recall that *PEA\** only closes a node if all of its successors have an $f$-value of no more than the optimal solution cost; if the span to the lower bound is small, each node can have at least one "bad" successor that exceeds this difference.

*IDDP* reduces the memory requirements further by a factor of about 6. The diagram also shows the maximum size of the *Open* list alone. For few sequences, the difference between the two is dominated by the linear length to store the solution path. As the problem size increases, however, the proportion of the *Closed* list of the total memory drops to about only 12 percent for 12 sequences. The total number of expansions (including all search stages) is slightly higher than in *PEA\**; however, due to optimizations made possible by the control of the expansion order, the execution time at 12 sequences is reduced by about a third.

|   | Num Exp | Time [sec] | Max Open | Max Open + Closed |
|---|---|---|---|---|
| | | | | |
| | | $A^*$ | | |
| 4 | 626 | 0.01 | 7805 | 8432 |
| 5 | 1599 | 0.05 | 32178 | 33778 |
| 6 | 3267 | 0.25 | 124541 | 127809 |
| 7 | 10781 | 1.94 | 666098 | 676880 |
| 8 | 116261 | 49.32 | 9314734 | 9430996 |
| 9 | 246955 | 318.58 | 35869671 | 36116627 |
| | | | | |
| | | $PEA^*$ | | |
| 3 | 448 | 0.01 | 442 | 442 |
| 4 | 716 | 0.01 | 626 | 626 |
| 5 | 2610 | 0.05 | 1598 | 1598 |
| 6 | 6304 | 0.33 | 3328 | 3331 |
| 7 | 23270 | 2.63 | 10874 | 10874 |
| 8 | 330946 | 87.24 | 118277 | 118277 |
| 9 | 780399 | 457.98 | 249279 | 249279 |
| 10 | 5453418 | 7203.17 | 1569815 | 1569815 |
| 11 | 20887627 | 62173.78 | 5620926 | 5620926 |
| 12 | 36078736 | 237640.14 | 9265949 | 9265949 |
| | | | | |
| | | $IDDP$ | | |
| 3 | 496 | 0.01 | 4 | 434 |
| 4 | 1367 | 0.02 | 9 | 443 |
| 5 | 6776 | 0.14 | 171 | 501 |
| 6 | 12770 | 0.59 | 414 | 972 |
| 7 | 26026 | 2.46 | 889 | 1749 |
| 8 | 362779 | 73.62 | 13620 | 19512 |
| 9 | 570898 | 250.48 | 21506 | 30009 |
| 10 | 4419297 | 4101.96 | 160240 | 192395 |
| 11 | 21774869 | 43708.14 | 860880 | 997163 |
| 12 | 36202456 | 158987.80 | 1417151 | 1616480 |

Table 1: Algorithm comparison for varying number of input sequences (elongation factors EF-TU and EF-1$\alpha$).

Since $PEA^*$ does not prune edges, its maximum space usage is always the total number of edges with $f$-value smaller than $g^*(t)$ (call these edges the *relevant edges*, since they have to be inspected by each admissible algorithm). In $IDDP$, on the other hand, the *Open* list can only comprise $k$ adjacent levels out of those edges (not counting the possible threshold overshoot, which would contribute a factor of at most 2). Thus, the improvement of $IDDP$ over $PEA^*$ will tend to increase with the overall number of levels (which is the sum of all string lengths), divided by the number of sequences; in other words, with the average sequence length.

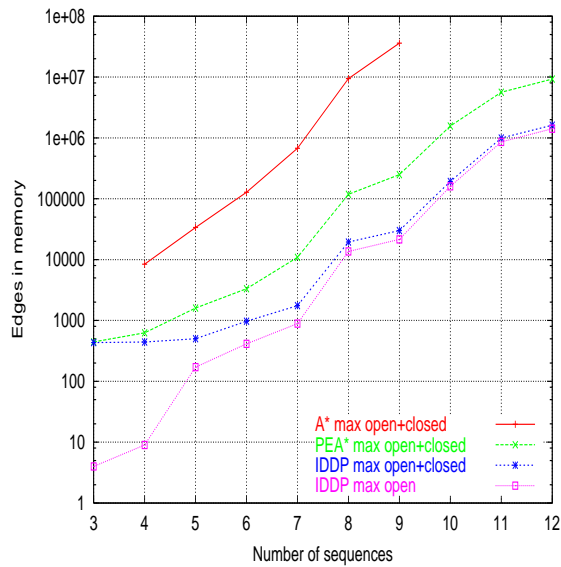

Figure 13: Memory requirements for $A^*$, $IDDP$, and $PEA^*$(elongation factors EF-TU and EF-1$\alpha$).

Moreover, the ratio depends on how well the heuristic suits the particular problem. Fig. 14 shows the distribution of all edges with $f$ value smaller or equal to $g^*(t)$, for the case of 9 of the example sequences. This problem is quite extreme as the bulk of these edges is concentrated in a small level band between 1050 and 1150. As an example with a more even distribution, Fig. 15 depicts the situation for problem `1cpt` from Reference 1 in the benchmark set $BAliBASE$ (Thompson et al., 1999) with heuristic $h_{all,3}$. In this case, the proportion of the overall 19492675 relevant edges that are maximal among all 4 adjacent levels amounts to only 0.2 percent. The maximum $Open$ size in $IDDP$ is 7196, while the total number of edges generated by $PEA^*$ is 327259, an improvement by about a factor of 45.

## 7.2 Multidimensional Heuristics

On the same set of sequences, we compared different improved heuristics in order to get an impression for their respective potential. Specifically, we ran $IDDP$ with heuristics $h_{pair}$, $h_{all,3}$, $h_{all,4}$, and $h_{one,k/2}$ at various thresholds $\delta$. Fig. 16 shows the total execution time for computing the heuristics, and performing the main search. In each case, we manually selected a value for $\delta$ which minimized this time. It can be seen that the times for $h_{one,k/2}$ lie only a little bit below $h_{pair}$; For few sequences (less than six), the computation of the heuristics $h_{all,3}$ and $h_{all,4}$ dominates their overall time. With increasing dimensions, how-ever, this investment starts to yield growing returns, with $h_{all,3}$ being the fastest algorithm, requiring only 5 percent of the time of $h_{pair}$ at 12 sequences.

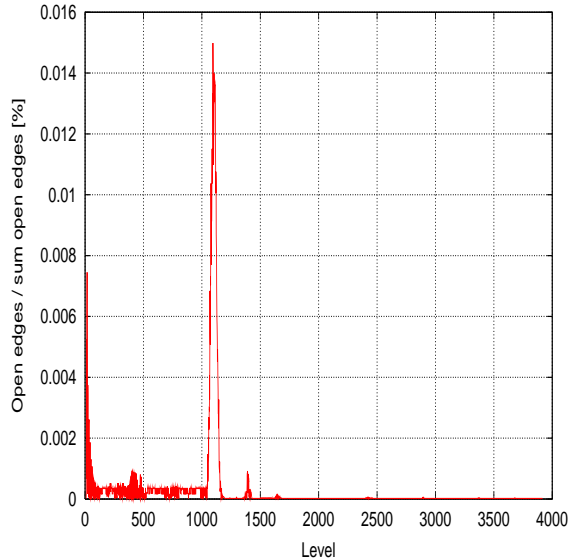

Figure 14: Distribution of relevant edges over levels (elongation factors EF-TU and EF-1$\alpha$); compare to the schematic projection in Fig. 4.

As far as memory is concerned, Fig. 17 reveals that the maximum size of the *Open* and *Closed* list, for the chosen $\delta$ values, is very similar for $h_{pair}$ and $h_{one,k/2}$ on the one hand, and $h_{all,3}$ and $h_{all,4}$ on the other hand.

At 12 sequences, $h_{one,6}$ saves only about 60 percent of edges, while $h_{all,3}$ only needs 2.6 percent and $h_{all,4}$ only 0.4 percent of the space required by the pairwise heuristic. Using *IDDP*, we never ran out of main memory; even larger test sets could be aligned, the range of the shown diagrams was limited by our patience to wait for the results for more than two days.

Based on the experienced burden of computing the heuristic, Kobayashi and Imai (1998) concluded that $h_{one,m}$ should be preferred to $h_{all,m}$. We do not quite agree with this judgment. We see that the heuristic $h_{all,m}$ is able to reduce the search space of the main search considerably stronger than $h_{one,m}$, so that it can be more beneficial with an appropriate amount of heuristic computation.

### 7.3 The Benchmark Database *BAliBASE*

*BAliBASE* (Thompson et al., 1999) is a widely used database of manually-refined multiple sequence alignments specifically designed for the evaluation and comparison of multiple sequence alignment programs. The alignments are classified into 8 reference sets. Reference 1 contains alignments of up to six about equidistant sequences. All the sequences are of sim-ilar length; they are grouped into 9 classes, indexed by sequence length and the percentage of identical amino acids in the same columns. Note that many of these problems are indeed much harder than the elongation factor examples from the previous section; despite consisting of fewer sequences, their dissimilarities are much more pronounced.

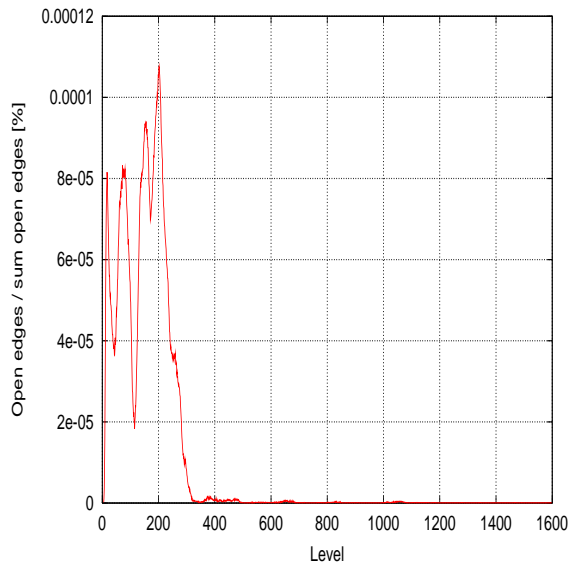

Figure 15: Distribution of relevant edges over levels, problem `1cpt` from *BAliBASE*.

We applied our algorithm to Reference 1, with substitution costs according to the PET91 matrix (Jones et al., 1992) and affine gap costs of $9 \cdot x + 8$, except for leading and trailing gaps, where no gap opening penalty was charged. For all instances, we precomputed the pairwise sub-alignments up to a fixed bound of 300 above the optimal solution; the optimal solution was found within this bound in all cases, and the effort is generally marginal compared to the overall computation. For all problems involving more than three sequences, the heuristic $h_{all,3}$ was applied.

Out of the 82 alignment problems in Reference 1, our algorithm could solve all but 2 problems (namely, *1pamA* and *gal4*) on our computer. Detailed results are listed in Tables 2 through 10.

Thompson, Plewniak, and Poch (1999) compared a number of widely used heuristic alignment tools using the so-called *SP*-score; their software calculates the percentage of correctly aligned pairs within the biologically significant motifs. They found that all programs perform about equally well for the sequences with medium and high amino acid identity; differences only occurred for the case of the more distant sequences with less than 25 percent identity, the so-called "twilight zone". Particularly challenging was the group of short sequences. In this subgroup, the three highest scoring programs are PRRP, CLUSTALX, and SAGA, with respective median scores of 0.560, 0.687, and 0.529. The medium score for the alignments found in our experiments amounts to 0.558; hence, it is about as good as PRRP, and only beaten by CLUSTALX. While we focused in our exper-iments on algorithmic feasibility rather than on solution quality, it would be worthwhile to attempt to improve the alignments found by these program using their more refined penalty functions. CLUSTALX, for example, uses different PAM matrices depending on the evolutionary distance of sequences; moreover, it assigns weights to sequences (based on a phylogenetic tree), and gap penalties are made position-specific. All of these improvements can be easily integrated into the basic sum-of-pairs cost function, so that we could attempt to compute an optimal alignment with respect to these metrics. We leave this line of research for future work.

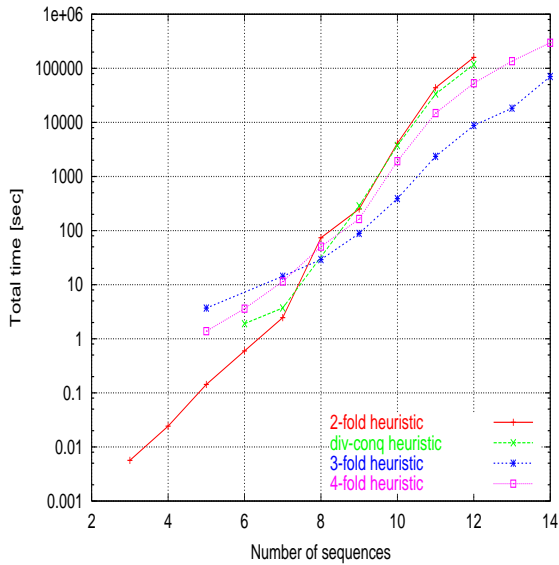

Figure 16: Comparison of execution times (including calculation of heuristics), elongation factors EF-TU and EF-1$\alpha$.

Fig. 18 shows the maximum number of edges that have to be stored in *Open* during the search, in dependence of the search threshold in the final iteration. For better comparability, we only included those problems in the diagram that consist of 5 sequences. The logarithmic scale emphasizes that the growth fits an exponential curve quite well. Roughly speaking, an increase of the cost threshold by 50 leads to a ten-fold increase in the space requirements. This relation is similarly applicable to the number of expansions (Fig. 19).

Fig. 20 depicts the proportion between the maximum *Open* list size and the combined maximum size of *Open* and *Closed*. It is clearly visible that due to the pruning of edges outside of possible solution paths, the *Closed* list contributes less and less to the overall space requirements the more difficult the problems become.

Finally, we estimate the reduction in the size of the *Open* list compared to all relevant edges by the ratio of the maximum *Open* size in the last iteration of *IDDP* to the total number of expansions in this stage, which is equal to the number of edges with $f$-value less or equal to the threshold. Considering possible overshoot of *IDDP*, algorithm *PEA\** would expand at least half of these nodes. The proportion ranges between 0.5 to 5 percent (cf. Fig. 21). Its considerable scatter indicates the dependence on individual problem properties; however, a slight average decrease can be noticed for the more difficult problems.

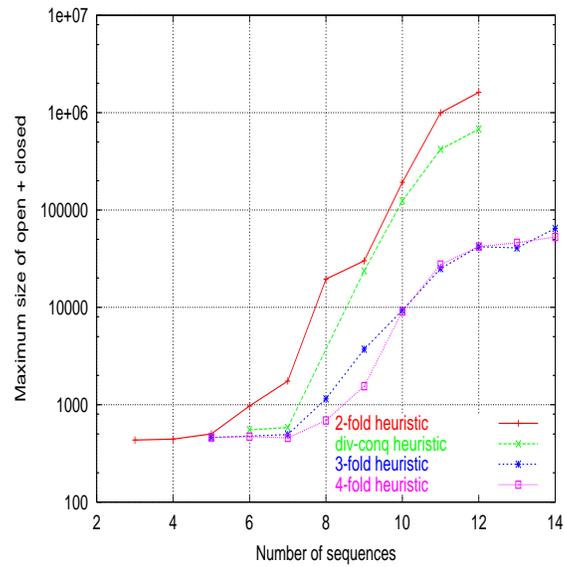

Figure 17: Combined maximum size of *Open* and *Closed*, for different heuristics (elongation factors EF-TU and EF-1$\alpha$).

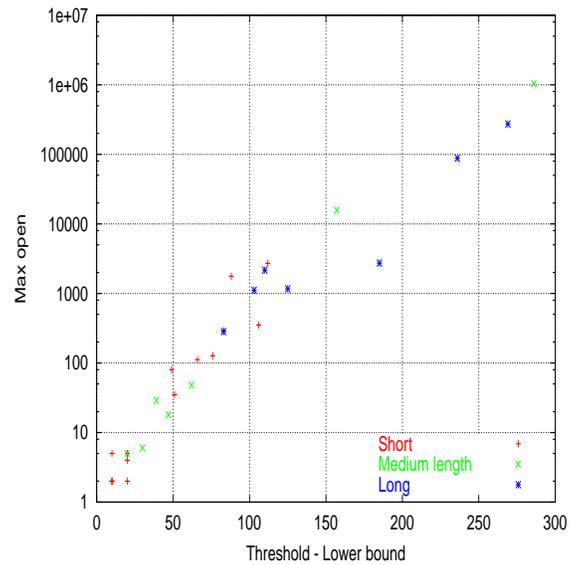

Figure 18: Maximum size of *Open* list, dependent on the final search threshold (*BAliBASE*).

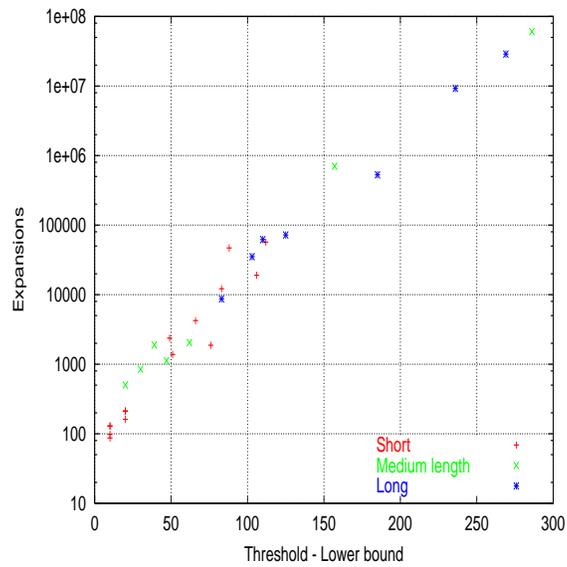

Figure 19: Number of expansions in the final search iteration (*BAliBASE*).

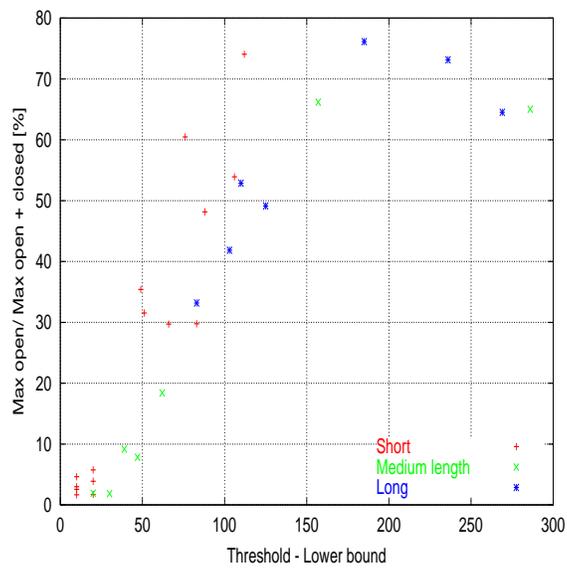

Figure 20: Maximum number of *Open* edges, divided by combined maximum of *Open* and *Closed* (*BAliBASE*).

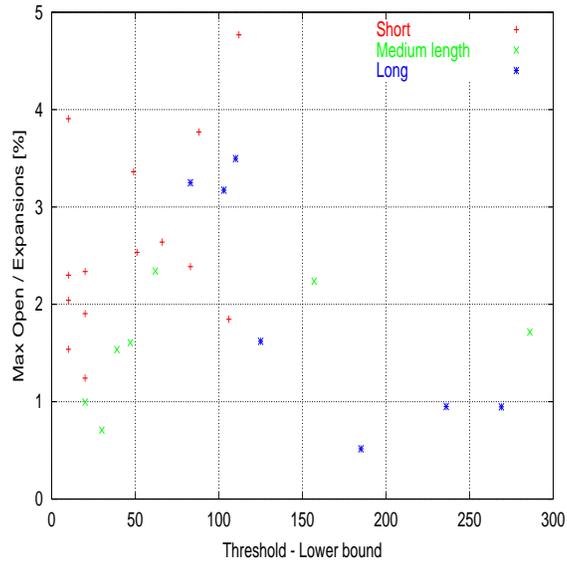

Figure 21: Percentage of reduction in *Open* size (*BAliBASE*).

## 8. Conclusion and Discussion

We have presented a new search algorithm for optimal multiple sequence alignment that combines the effective use of a heuristic bound as in best-first search with the ability of the dynamic programming approach to reduce the maximum size of the *Open* and *Closed* lists by up to one order of magnitude of the sequence length. The algorithm performs a series of searches with successively increasing bounds that explore the search space in *DP* order; the thresholds are chosen adaptively so that the expected overhead in recomputations is bounded by a constant factor.

We have demonstrated that the algorithm can outperform one of the currently most successful algorithms for optimal multiple sequence alignments, *Partial Expansion A\**, both in terms of computation time and memory consumption. Moreover, the iterative-deepening strategy alleviates the use of partially computed higher-dimensional heuristics. To the best of our knowledge, the algorithm is the first one that is able to solve standard benchmark alignment problems in *BAliBASE* with a biologically realistic cost function including affine gap costs without end gap penalties. The quality of the alignment is in the range of the best heuristic programs; while we have concentrated on algorithmic feasibility, we deem it worthwhile to incorporate their refined cost metrics for better results; we will study this question in future work.

Recently, we learned about related approaches developed simultaneously and independently by Zhou and Hansen (2003b, 2004). *SweepA\** explores a search graph according to layers in a partial order, but still uses the *f*-value for selecting nodes within one layer. *Breadth-First Heuristic Search* implicitly defines the layers in a graph with uniform costs according to the breadth-first traversal. Both algorithms incorporate upper bounds on the optimal solution cost for pruning; the idea of adaptive threshold determination to limit re-expansion overhead to a constant factor is not described. Moreover, they do not consider the flexible use of additional memory to minimize the divide-and-conquer solution reconstruction phase.

Although we described our algorithm entirely within the framework of the MSA problem, it is straightforward to transfer it to any domain in which the state space graph is directed and acyclic. Natural candidates include applications where such an ordering is imposed by time or space coordinates, e.g., finding the most likely path in a Markov model.

Two of the *BAliBASE* benchmark problems could still not be solved by our algorithm within the computer's main memory limit. Future work will include the integration of techniques exploiting secondary memory. We expect that the level-wise exploration scheme of our algorithm lends itself naturally to *external search algorithms*, another currently very active research topic in Artificial Intelligence and theoretical computer science.

## Acknowledgments

The author would like to thank the reviewers of this article whose comments have helped in significantly improving it.

## Appendix A

Table 2: Results for *BAliBASE* Reference 1, group of short sequences with low amino acid identity. The columns denote: S — number of aligned sequences; $\delta$ — upper bound for precomputing optimal solutions for partial problems in last iteration of main search; $g^*(t)$ — optimal solution cost; $h(s)$ — lower bound for solution cost, using heuristics; #Exp — total number of expansions in all iterations of the main search; #Op — peak number of edges in *Open* list over the course of the search; #Op+Cl — peak combined number of edges in either *Open* or *Closed* list during search; #Heu — peak number of sub—alignment edge costs stored as heuristic; Time: total running time including auxiliary and main search, in seconds; Mem — peak total memory usage for face alignments, heuristic, and main search, in KB.

|        | S | $\delta$ | $g^*(t)$ | $h(s)$ | #Exp    | #Op    | #Op+Cl | #Heu    | Time    | Mem   |
|--------|---|----------|----------|--------|---------|--------|--------|---------|---------|-------|
| 1aboA  | 5 | 57       | 9006     | 8898   | 3413786 | 104613 | 176126 | 1654547 | 331.029 | 15568 |
| 1idy   | 5 | 50       | 8165     | 8075   | 1732008 | 74865  | 121404 | 970933  | 167.867 | 10893 |
| 1r69   | 4 | 20       | 6215     | 6183   | 634844  | 19938  | 41719  | 88802   | 22.517  | 3568  |
| 1tvxA  | 4 | 44       | 5532     | 5488   | 1263849 | 24226  | 48633  | 476622  | 52.860  | 5278  |
| 1ubi   | 4 | 30       | 7395     | 7357   | 1614286 | 26315  | 54059  | 289599  | 62.133  | 5448  |
| 1wit   | 5 | 69       | 14287    | 14176  | 6231378 | 209061 | 351582 | 2442098 | 578.907 | 27273 |
| 2trx   | 4 | 20       | 7918     | 7899   | 63692   | 3502   | 5790   | 127490  | 4.572   | 1861  |

Table 3: Short sequences, medium similarity.

|       | S | $\delta$ | $g^*(t)$ | $h(s)$ | #Exp    | #Op   | #Op+Cl | #Heu   | Time    | Mem  |
|-------|---|----------|----------|--------|---------|-------|--------|--------|---------|------|
| 1aab  | 4 | 20       | 6002     | 5984   | 263     | 12    | 83     | 4404   | 0.572   | 691  |
| 1fjlA | 6 | 20       | 13673    | 13625  | 900     | 106   | 155    | 19573  | 0.985   | 1589 |
| 1hfh  | 5 | 30       | 16556    | 16504  | 137914  | 4852  | 8465   | 70471  | 14.077  | 2882 |
| 1hpi  | 4 | 20       | 5858     | 5835   | 1560    | 83    | 164    | 5269   | 0.679   | 656  |
| 1csy  | 5 | 30       | 14077    | 14026  | 52718   | 3872  | 5613   | 56191  | 6.165   | 2252 |
| 1pfc  | 5 | 30       | 15341    | 15277  | 118543  | 6477  | 8905   | 55887  | 11.850  | 2478 |
| 1tgxA | 4 | 20       | 4891     | 4856   | 18987   | 543   | 1080   | 5507   | 1.196   | 649  |
| 1ycc  | 4 | 20       | 8926     | 8903   | 54049   | 1118  | 2010   | 77156  | 3.780   | 1644 |
| 3cyr  | 4 | 48       | 8480     | 8431   | 583260  | 13422 | 25806  | 193690 | 22.592  | 3076 |
| 451c  | 5 | 49       | 11440    | 11333  | 1213162 | 38004 | 54115  | 583363 | 111.675 | 6529 |

Table 4: Short sequences, high similarity.

|       | S | $\delta$ | $g^*(t)$ | $h(s)$ | #Exp  | #Op  | #Op+Cl | #Heu  | Time  | Mem  |
|-------|---|----------|----------|--------|-------|------|--------|-------|-------|------|
| 1aho  | 5 | 20       | 8251     | 8187   | 30200 | 2255 | 3074   | 10971 | 3.175 | 1042 |
| 1csp  | 5 | 20       | 8434     | 8427   | 90    | 2    | 78     | 3528  | 0.569 | 784  |
| 1dox  | 4 | 20       | 7416     | 7405   | 782   | 50   | 186    | 8406  | 0.652 | 823  |
| 1fkj  | 5 | 20       | 13554    | 13515  | 2621  | 140  | 222    | 10925 | 0.945 | 1511 |
| 1fmb  | 4 | 20       | 7571     | 7568   | 172   | 4    | 108    | 1804  | 0.540 | 788  |
| 1krn  | 5 | 20       | 9752     | 9747   | 101   | 1    | 87     | 6244  | 0.623 | 1035 |
| 1plc  | 5 | 20       | 12177    | 12152  | 454   | 25   | 103    | 10641 | 0.728 | 1415 |
| 2fxb  | 5 | 20       | 6950     | 6950   | 88    | 2    | 71     | 1432  | 0.534 | 617  |
| 2mhr  | 5 | 20       | 14317    | 14306  | 256   | 4    | 121    | 7853  | 0.668 | 1558 |
| 9rnt  | 5 | 20       | 12382    | 12367  | 350   | 19   | 108    | 6100  | 0.695 | 1250 |

Table 5: Medium-length sequences, low similarity.

|        | S | $\delta$ | $g^*(t)$ | $h(s)$ | #Exp       | #Op      | #Op+Cl   | #Heu     | Time        | Mem    |
|--------|---|----------|----------|--------|------------|----------|----------|----------|-------------|--------|
| 1bbt3  | 5 | 160      | 30598    | 30277  | 902725789  | 11134608 | 15739188 | 23821767 | 43860.175   | 927735 |
| 1sbp   | 5 | 200      | 42925    | 42512  | 2144000052 | 6839269  | 11882990 | 65341855 | 106907.000  | 735053 |
| 1havA  | 5 | 200      | 31600    | 31234  | 2488806444 | 10891271 | 16321376 | 58639851 | 132576.000  | 927735 |
| 1uky   | 4 | 94       | 18046    | 17915  | 179802791  | 659435   | 1281339  | 15233338 | 7006.560    | 106184 |
| 2hsdA  | 4 | 96       | 21707    | 21604  | 65580608   | 293357   | 668926   | 12497761 | 2646.880    | 67788  |
| 2pia   | 4 | 161      | 22755    | 22616  | 97669470   | 789446   | 1673807  | 25718770 | 4310.030    | 142318 |
| 3grs   | 4 | 126      | 20222    | 20061  | 107682032  | 640391   | 1396982  | 24104710 | 4267.880    | 130425 |
| kinase | 5 | 200      | 45985    | 45520  | 2446667393 | 13931051 | 19688961 | 32422084 | 125170.460  | 927734 |

Table 6: Medium-length sequences, medium similarity.

|        | S | $\delta$ | $g^*(t)$ | $h(s)$ | #Exp     | #Op     | #Op+Cl  | #Heu     | Time     | Mem    |
|--------|---|----------|----------|--------|----------|---------|---------|----------|----------|--------|
| 1ad2   | 4 | 20       | 16852    | 16843  | 379      | 16      | 221     | 27887    | 0.959    | 2186   |
| 1aym3  | 4 | 20       | 19007    | 18978  | 466536   | 4801    | 8914    | 83634    | 15.386   | 3163   |
| 1gdoA  | 4 | 58       | 20696    | 20613  | 10795040 | 57110   | 102615  | 1265777  | 363.549  | 12028  |
| 1ldg   | 4 | 20       | 25764    | 25736  | 446123   | 4981    | 9052    | 169038   | 16.115   | 4484   |
| 1mrj   | 4 | 20       | 20790    | 20751  | 252601   | 4067    | 7380    | 33942    | 8.694    | 2905   |
| 1pgtA  | 4 | 50       | 17442    | 17398  | 1870204  | 19200   | 32476   | 485947   | 73.066   | 5869   |
| 1pii   | 4 | 20       | 20837    | 20825  | 25256    | 584     | 1414    | 116670   | 3.089    | 3338   |
| 1ton   | 5 | 102      | 32564    | 32428  | 13571887 | 351174  | 526102  | 11549908 | 1373.180 | 58704  |
| 2cba   | 5 | 160      | 40196    | 39914  | 60545205 | 1037828 | 1595955 | 19186631 | 2904.651 | 140712 |

Table 7: Medium-length sequences, high similarity.

| | S | δ | g*(t) | h(s) | #Exp | #Op | #Op+Cl | #Heu | Time | Mem |
|---|---|---|---|---|---|---|---|---|---|---|
| 1amk | 5 | 20 | 31473 | 31453 | 447 | 7 | 259 | 13120 | 0.825 | 3366 |
| 1ar5A | 4 | 20 | 15209 | 15186 | 3985 | 128 | 356 | 22220 | 1.066 | 1755 |
| 1ezm | 5 | 20 | 37396 | 37381 | 613 | 4 | 324 | 15751 | 0.836 | 3900 |
| 1led | 4 | 20 | 18795 | 18760 | 93220 | 2956 | 4951 | 39962 | 3.761 | 2564 |
| 1ppn | 5 | 20 | 27203 | 27159 | 18517 | 489 | 864 | 20209 | 2.545 | 2991 |
| 1pysA | 4 | 20 | 19242 | 19215 | 10810 | 190 | 801 | 14344 | 1.200 | 2224 |
| 1thm | 4 | 20 | 21470 | 21460 | 361 | 2 | 293 | 8090 | 0.682 | 2469 |
| 1tis | 5 | 20 | 35444 | 35395 | 31996 | 448 | 915 | 42716 | 4.409 | 4122 |
| 1zin | 4 | 20 | 16562 | 16546 | 771 | 23 | 225 | 6619 | 0.654 | 1767 |
| 5ptp | 5 | 20 | 29776 | 29735 | 6558 | 309 | 539 | 37883 | 1.767 | 3600 |

Table 8: Long sequences, low similarity.

| | S | δ | g*(t) | h(s) | #Exp | #Op | #Op+Cl | #Heu | Time | Mem |
|---|---|---|---|---|---|---|---|---|---|---|
| 1ajsA | 4 | 160 | 38382 | 38173 | 318460012 | 1126697 | 2310632 | 27102589 | 9827.233 | 208951 |
| 1cpt | 4 | 160 | 39745 | 39628 | 873548 | 5260 | 12954 | 10494564 | 223.926 | 32119 |
| 1lvl | 4 | 160 | 43997 | 43775 | 537914936 | 1335670 | 2706940 | 37491416 | 16473.420 | 255123 |
| 1ped | 3 | 50 | 15351 | 15207 | 2566052 | 7986 | 27718 | 0 | 20.035 | 4447 |
| 2myr | 4 | 200 | 43414 | 43084 | 3740017645 | 7596730 | 45488908 | 118747184 | 136874.980 | 927735 |
| 4enl | 3 | 50 | 16146 | 16011 | 5169296 | 9650 | 30991 | 0 | 41.716 | 5589 |

Table 9: Long sequences, medium similarity.

| | S | δ | g*(t) | h(s) | #Exp | #Op | #Op+Cl | #Heu | Time | Mem |
|---|---|---|---|---|---|---|---|---|---|---|
| 1ac5 | 4 | 92 | 37147 | 37020 | 169779871 | 732333 | 1513853 | 18464119 | 6815.760 | 124877 |
| 1adj | 4 | 20 | 32815 | 32785 | 207072 | 3106 | 5145 | 96176 | 7.829 | 4595 |
| 1bgl | 4 | 243 | 78366 | 78215 | 188429118 | 857008 | 1744149 | 101816849 | 8795.000 | 291618 |
| 1dlc | 4 | 106 | 47430 | 47337 | 14993317 | 65288 | 126608 | 12801019 | 843.402 | 43158 |
| 1eft | 4 | 56 | 31377 | 31301 | 9379999 | 42620 | 72502 | 1476154 | 334.475 | 13115 |
| 1fieA | 4 | 86 | 53321 | 53241 | 6905957 | 46779 | 90937 | 6040375 | 348.134 | 26884 |
| 1gowA | 4 | 166 | 38784 | 38632 | 45590739 | 275256 | 544800 | 31318364 | 2251.190 | 99537 |
| 1pkm | 4 | 89 | 36356 | 36256 | 11197890 | 75144 | 140472 | 5962640 | 505.778 | 27244 |
| 1sesA | 4 | 58 | 57670 | 57557 | 4755983 | 96014 | 136677 | 3585721 | 463.962 | 27452 |
| 2ack | 5 | 250 | 76937 | 76466 | 994225856 | 8077412 | 12436928 | 75819994 | 32965.522 | 765715 |
| arp | 5 | 143 | 54939 | 54696 | 182635167 | 1291185 | 2160263 | 38368530 | 15972.000 | 193364 |
| glg | 5 | 160 | 74282 | 74059 | 9251905 | 87916 | 120180 | 22622910 | 733.202 | 72148 |

Table 10: Long sequences, high similarity.

|  | S | δ | $g^*(t)$ | $h(s)$ | #Exp | #Op | #Op+Cl | #Heu | Time | Mem |
|---|---|---|---|---|---|---|---|---|---|---|
| 1ad3 | 4 | 20 | 33641 | 33604 | 104627 | 2218 | 3461 | 34539 | 4.196 | 3968 |
| 1gpb | 5 | 54 | 101296 | 101231 | 1232707 | 62184 | 98476 | 2702949 | 178.610 | 25698 |
| 1gtr | 5 | 60 | 55242 | 55133 | 2037633 | 54496 | 91656 | 1916127 | 226.791 | 18050 |
| 1lcf | 6 | 160 | 149249 | 148854 | 181810148 | 3235312 | 3824010 | 28614215 | 15363.051 | 294688 |
| 1rthA | 5 | 128 | 69296 | 69133 | 14891538 | 71081 | 105082 | 24587882 | 1721.070 | 70569 |
| 1taq | 5 | 250 | 133723 | 133321 | 1693501628 | 9384718 | 17298456 | 145223167 | 5713.240 | 1170673 |
| 3pmg | 4 | 51 | 42193 | 42133 | 1036943 | 8511 | 15540 | 777639 | 50.796 | 8133 |
| actin | 5 | 53 | 48924 | 48826 | 824295 | 35283 | 53009 | 777058 | 96.147 | 11198 |